\documentclass[11pt]{article}
\usepackage[margin=1in]{geometry}
\usepackage[T1]{fontenc}
\usepackage[utf8]{inputenc}
\usepackage{graphicx}
\usepackage{float}
\usepackage{hyperref}
\usepackage{pdfpages}
\title{Evaluating LLMs for Answering Student Questions in Introductory Programming Courses}
\author{Thomas Van Mullem, Bart Mesuere, Peter Dawyndt}

\date{}

\begin{document}

\maketitle

\begin{abstract}
The rapid emergence of Large Language Models (LLMs) presents both opportunities and challenges for programming education. While students increasingly use generative AI tools, direct access often hinders the learning process by providing complete solutions rather than pedagogical hints. Concurrently, educators face significant workload and scalability challenges when providing timely, personalized feedback. This study investigates the capabilities of LLMs to safely and effectively assist educators in answering student questions within a CS1 programming course. To achieve this, we established a rigorous, reproducible evaluation process by curating a benchmark dataset of 170 authentic student questions from a learning management system, paired with ground-truth responses authored by subject matter experts. Because traditional text-matching metrics are insufficient for evaluating open-ended educational responses, we developed and validated a custom LLM-as-a-Judge metric optimized for assessing pedagogical accuracy. Our findings demonstrate that models, such as Gemini 3 flash, can surpass the quality baseline of typical educator responses, achieving high alignment with expert pedagogical standards. To mitigate persistent risks like hallucination and ensure alignment with course-specific context, we advocate for a "teacher-in-the-loop" implementation. Finally, we abstract our methodology into a task-agnostic evaluation framework, advocating for a shift in the development of educational LLM tools from ad-hoc, post-deployment testing to a quantifiable, pre-deployment validation process.
\end{abstract}

\section{Keywords:}
computer science, programming, LLM, LLM-as-a-Judge, Q\&A, task-agnostic evaluation framework, pre-deployment evaluation

\section{Introduction}
The emergence of Large Language Models (LLMs) capable of solving complex programming tasks has disrupted programming education. Recent studies show that current models can solve nearly all standard CS1 and CS2 exercises with high accuracy \cite{finnie-ansley_robots_2022,finnie-ansley_my_2023,reeves_evaluating_2023,savelka_thrilled_2023}. Consequently, generative AI tools have become increasingly relevant in students’ careers \cite{prather_robots_2023}, with students utilizing them to generate code, explain concepts, and debug errors. However, while these tools provide quick, detailed responses, they frequently provide complete solutions \cite{hellas_exploring_2023}, thereby hindering the learning process instead of supporting it. For instance, \cite{kazemitabaar_how_2024} showed the most common approach used by students is a single prompt to generate the solution. This approach resulted in consistent negative student performance on post-test evaluation scores, indicating that unrestricted LLM access is pedagogically undesirable. Conversely, the same authors \cite{kazemitabaar_how_2024} also highlighted that a hybrid approach, where students manually write parts of the code while using AI to generate other parts, yielded positive trends. This suggests that LLMs can offer valuable learning support, but the risk of student over-reliance underscores the critical importance of good instruction and educator oversight.

As these tools are here to stay, it is important to explore their pedagogical opportunities \cite{becker_programming_2023,kasneci_chatgpt_2023}. While recent studies have focused on investigating model capabilities for generating solutions to programming exercises \cite{finnie-ansley_robots_2022,finnie-ansley_my_2023,reeves_evaluating_2023,savelka_thrilled_2023}, generating personalized assignments \cite{jacobs_unlimited_2025, sarsa_automatic_2022} and explanations for students \cite{leinonen_using_2023, sarsa_automatic_2022}, students still rely on educators to answer questions, give feedback and solve problems they encounter. This reliance poses a challenge regarding scalability and educator workload, because providing accurate, comprehensive and personalized responses to student questions takes considerable time. To address this, automated tools and chatbots like CodeHelp \cite{liffiton_codehelp_2024} and CodeAid \cite{kazemitabaar_codeaid_2024} have been developed to provide immediate, automated assistance. However, these systems provide LLM-generated answers directly to students, thereby exposing students to potential hallucinations, incorrect information or pedagogically unsound issues (complete solutions instead of hints, use of concepts that students have not grasped, use of terminology/strategy that differs from what teachers apply). If a student receives erroneous information from these institutionally provided tools, we risk eroding student trust. If this trust is lost, students will likely turn towards general-purpose models like ChatGPT or Gemini, bypassing pedagogical guardrails entirely. This shows the need for a way to provide students with scalable, educator-verified answers. 

Furthermore, the adoption of LLM-powered educational tools has highlighted a critical methodological gap in current research: many of these educational tools are deployed directly into classroom settings without rigorous pre-evaluation of their generated responses. To ensure these systems are safe and effective, their output should be validated against ground truth data prior to real-world deployment. However, automatically evaluating open-ended text, such as answers to student questions, has historically proven difficult. Recent advancements like LLM-as-a-Judge \cite{zheng_judging_2023} offer a promising solution. In this approach, an LLM is instructed to give a score to a text based on provided criteria, allowing us to evaluate elements like factual correctness, similarity and semantic values of a text. Recent studies have shown that these judges can serve as viable replacements for human evaluation, as they can achieve high agreement rates with humans \cite{wang_can_2025}.

In this paper, we address the evaluation gap by leveraging the LLM-as-a-Judge technique as a metric to quantify the performance of LLMs executing the Q\&A task. We develop several prompt/model combinations designed to answer student questions in a CS1 programming course and evaluate their performance before classroom deployment. To achieve this, we create a benchmark consisting of authentic student questions as input data, ground truths and (automated) metrics. This benchmark gives us a scalable and reproducible way to decide if LLMs are fit for the task and what prompt/model combination is best. Furthermore, the benchmark provides us with a future-proof way of evaluating new prompts, models and technologies. In the ever-changing landscape of LLMs, having a reliable method to evaluate new models and prompts is indispensable. This leads to the following research questions:
\begin{description}
  \item[RQ1:] To what extent can LLMs generate pedagogically appropriate answers to student questions in a CS1 programming course?
  \item[RQ2:] How can we establish a reproducible, scientific process for developing and evaluating LLM-based (educational) tools?
  \item[RQ3:] What reusable workflow or set of principles can be distilled for designing and evaluating similar tools across domains and tasks?
\end{description}

\section{Methods}
The goal of this work is to assess the answer generation capabilities of LLMs to programming questions asked by students. We want to explore whether current state-of-the-art (SOTA) LLMs are good enough to answer student questions in CS1 courses, prior to their deployment in a classroom environment. In order to evaluate the performance of different prompts and SOTA models, we structured our research into four distinct phases. The first phase involves collecting and curating a representative input dataset, which comprises authentic student questions, corresponding code submissions, and other system outputs, from a dedicated learning management system. Following this, a reliable ground truth is established for the selected data by having subject matter experts (SMEs) author pedagogically sound and verified answers to the selected student questions. The third phase focuses on identifying success criteria and selecting or creating accompanying metrics. Finally, we create a robust evaluation framework based on the input data, ground truth and metrics. Using this framework, we systematically evaluate the performance of different actors performing the task. These actors can be prompt/model combinations or even outputs generated by educators. Each of these phases is detailed in the subsequent sections.

\subsection{Input data}
To investigate whether LLMs can effectively answer student questions in an introductory computer science classroom, a dataset was compiled from Dodona \cite{van_petegem_dodona_2023}, a dedicated learning management system (LMS) for learning to code. The input dataset comprises student questions from an introductory Python programming course (2023-2024) and includes both English (EN) and Dutch (NL) items. Students ask questions about the weekly exercises both online through Dodona and in person during on-campus lab sessions. Over the course of one semester 1140 questions were asked by students through the Q\&A module of the LMS, from which 200 questions were randomly sampled across the semester. Of these questions, 30 were left out due to privacy considerations (GDPR; personal information was included that should not be sent to LLM providers) or poor quality (e.g., lack of sufficient context, unrelated questions, questions about grades). The final input dataset contains 170 questions, of which 134 have an answer provided by one of the educators of the course. The remaining questions were self-resolved, or marked by the student as no longer relevant. Each entry in the input dataset includes the student’s question and submitted code, the line number of the question, their preferred natural language (English or Dutch), the assignment description, details on failing test cases, linting errors and the programming language. A sample entry based on the “radians example” shown in Figure~\ref{fig:figure1} can be found in the supplementary material. The figure shows a student question at the top of the code, recognizable by the purple line in front of it. The educator’s response is provided as an annotation at a different line of the code.

\begin{figure}[H]
    \centering
    \includegraphics[width=\textwidth]{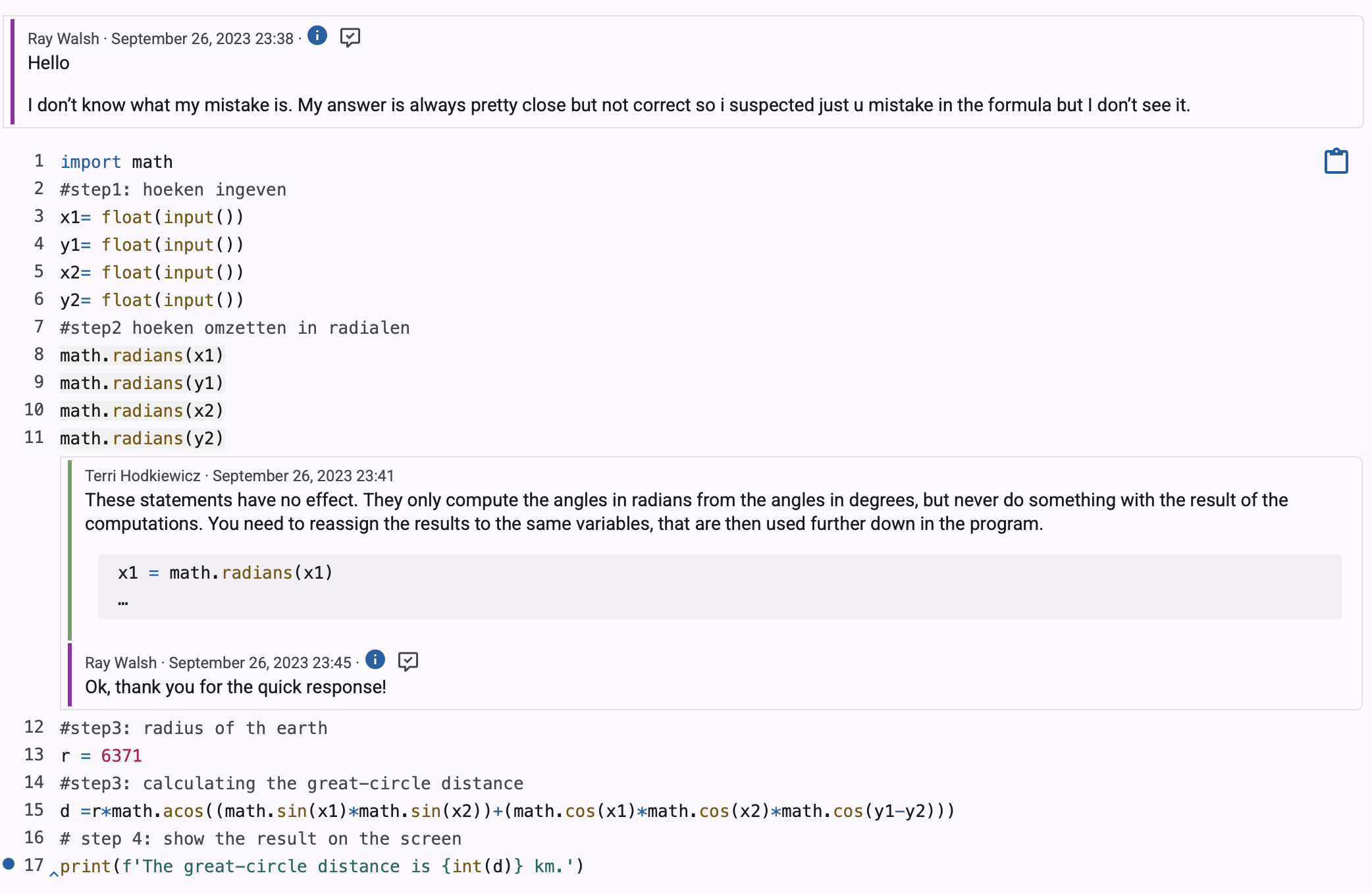}
    \caption{Example of a question asked by a student (Ray Walsh) at the top of the code, followed by the answer provided by one of the educators (Tim Hodkiewicz) at line 11 of the code. Names have been pseudonymized.}
    \label{fig:figure1}
\end{figure}

\subsection{Ground truth}
\begin{figure}[H]
    \centering
    \includegraphics[width=\textwidth]{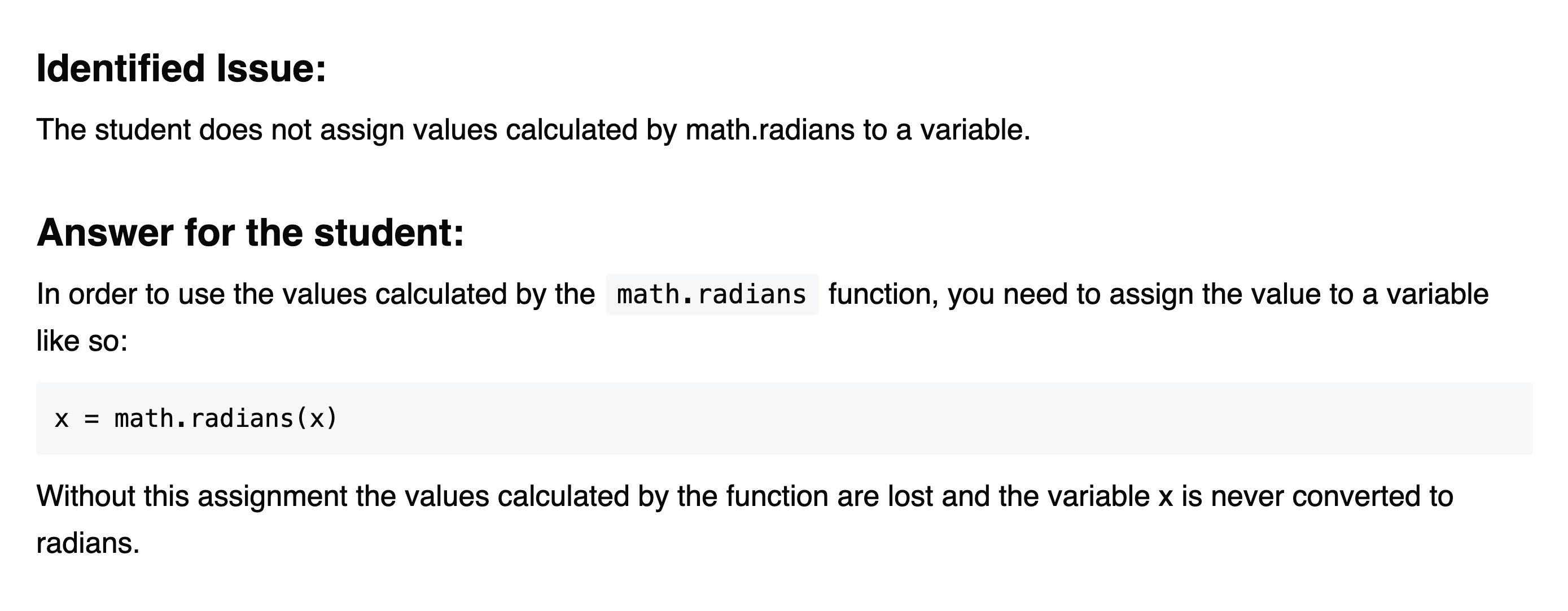}
    \caption{Ground truth compiled by a subject matter expert; both the identified issue and an answer for the student are provided. This answer is used as ground truth to the question from Figure~\ref{fig:figure1}.}
    \label{fig:figure2}
\end{figure}

To establish a reliable ground truth for evaluating task performance, an expert answered all 170 selected student questions from the input dataset. For questions directly tied to a student’s implementation, the expert verified the ground truth by applying the suggested strategy. General answers about coding style, errors or programming concepts adhere to best practices of the programming language (Python). The ground truths do not reveal full algorithms or complete code solutions. Each output consists of an identified issue, which provides a brief summary of the problem, and an ‘answer for the student’, which guides the student towards the correct solution by highlighting specific areas for reconsideration and specific hints. The ground truth for the radians example can be seen in Figure~\ref{fig:figure2}.

\subsection{Metrics}
In order to evaluate how different actors perform the Q\&A task, we defined several success criteria. Success criteria define what is considered a successful completion of the task at hand. For the Q\&A task, pedagogical accuracy is our primary criterion. Answers provided by an actor should contain the same teaching points as provided in the ground truth. However, the practical implementation also requires considering the LLM’s operational cost-efficiency. This resulted in two metrics for our research: the comparison of the ground truth with the actor’s output and the cost of answering a single question.

Answers to student questions are diverse and open-ended. They can vary in phrasing, structure, level of detail and they can provide different solutions to the same problem. Because of this variation, selecting a fitting metric is crucial. Traditional NLP metrics such as BLEU \cite{papineni_bleu_2001}, ROUGE \cite{lin_rouge_2004} or exact match are not suitable for evaluating the Q\&A task. These metrics heavily depend on surface-level similarity, which is unable to recognize the aforementioned characteristics of answers to student’s questions. Given these limitations, we successively explored LLM-as-a-Judge metrics. This approach utilizes an LLM to compare the actor's generated answer against the ground truth, scoring the output based on defined criteria like factual correctness and semantic similarity. We first investigated general-purpose LLM-as-a-Judge metrics including third-party libraries that provide LLM based evaluations such as context-precision, context-recall and G-Eval \cite{liu_g-eval_2023}. All of these metrics call an LLM with a custom prompt that is injected with the actor’s answer, ground truth and examples that show the LLM how to score certain responses. While this approach initially provided promising results, it lacked explainability and exhibited non-deterministic behavior. Key concepts like “precision” and “recall” lack concrete definitions when applied to the comparison of two texts. The concepts “false-positive”, “true-positive”, “false-negative” and “true-negative” in precision and recall are not easily defined for the comparison of an expert answer to a student’s question and an LLM-generated answer. 

As neither traditional metrics nor off-the-shelf LLM-as-a-Judge metrics met our needs, a custom LLM-as-a-Judge was developed. The development of this judge focused on identifying core teaching points rather than exact text matches. To ensure the judge’s scores are meaningful, we aligned them with scores of subject matter experts (SMEs). A dataset of 100 LLM-generated answers with ground truth values (scores assigned by an SME) was constructed and used to align the prompt with the expert's scoring. This dataset consists of 50 samples representing overly verbose, “filler heavy” explanations and 50 samples with concise but incomplete responses. SMEs manually annotated this dataset using a custom scoring rubric (0-5), establishing a benchmark for the LLM-as-a-Judge. These scores were then used to iteratively refine the judge’s prompt. By analyzing the scoring distribution through Cohen’s weighted kappa \cite{cohen_weighted_1968} and heatmap visualizations, the delta between the model’s and expert’ scores was minimized. This alignment process ensured that the judge was steered toward the same evaluation patterns as the SMEs, resulting in a metric that prioritizes pedagogical accuracy and completeness. This provides us with a scalable alternative to manual evaluation.

\subsection{Actor Evaluation}
The final step is actor evaluation. We gather data on different actors (referring to prompt/model combinations, humans, or a hybrid of both) performing the task to make an informed decision on who or what should execute it. The selection depends on the performance of these actors relative to a baseline. Before evaluating different models, we establish this baseline using the original answers given by the educators during the course. We term this the human baseline, which reflects real-world, time-constrained educator performance and is distinct from the expert-authored ground truth. This baseline contains the “Best Available Human” (BAH) responses. We evaluate these responses the same way as we do with LLM-generated answers by comparing them to the ground truth using the custom LLM-as-a-Judge metric. After establishing the BAH baseline, a cost-effective, contemporary model is used to perform prompt engineering. During our research, we used Gemini 2.5 flash to obtain a well-performing prompt, which is described in the next section. The resulting prompt can then be used to compare the performance of different models. Both an intra-family comparison and an inter-family comparison are performed. The intra-family comparison is performed on the Gemini family and shows how different generations of the same model perform. After this analysis, models from different families (OpenAI, Anthropic and Gemini) are tested to find the optimal prompt/model combination. Based on the BAH baseline and the performance of models, the best fitting actor can be selected.

\subsubsection{Prompt Engineering}
During prompt engineering, our primary goal was to develop an optimal instruction set for the LLM that performs the Q\&A task. This involves crafting a single, detailed prompt that combines all relevant student context (question, code, assignment details, etc.) with explicit instructions for the LLM. The prompt's design ensures the model generates pedagogically sound responses that align with our ground truth, ultimately maximizing the LLM's performance for the Q\&A task and providing a robust benchmark for model comparison.

The process involves systematically evaluating prompt variations against the ground truths using the custom LLM-as-a-Judge that was previously described. This is an iterative process where a wide range of prompts are researched via trial and error. It consists of selecting the right data, optimally formatting the data, wording and rewording desired outcomes, correcting for common errors and exploring new prompting techniques. Various prompts were tried, as small variations in global structure, lexical choices and grammatical structure can significantly influence the model’s performance. The most important observations can be grouped into three overarching categories. Data selection, data formatting and optimization.

Selecting the right input data is crucial, since this data provides context to the model, which may result in better performance. Introducing too much clutter can “confuse” the model and lead to diminishing results \cite{liu_lost_2024}. During prompt engineering, we applied a ‘leave-one-out’ approach on the available input data. The influence of every piece of data was studied by excluding parts of the input data and observing whether or not the performance dropped significantly.

The next prompt engineering step consists of formatting the input data appropriately. The only way to test this is by trial and error. The representation of the submitted code and student question are straightforward. These are unstructured texts that can be passed to the model as is. Assignments were converted to Markdown to ensure consistency and uniform formatting. Dodona returns software testing results in a complicated JSON structure with little restrictions, which requires some filtering before this data can be passed to the model. Subsequently, different orderings of the input data were explored and different ways of splitting up the content were evaluated. XML-tags proved to be the best delimiters and the order of the information did not have a significant impact on the performance of the models.

The third prompt engineering step is optimization; different prompts are iteratively tested and adjusted in order to improve the quality, consistency and accuracy of the generated answers. This involves carefully rewording parts of the prompt to get the desired behavior, eliminating unintended behaviors and getting rid of common mistakes. In particular, it was found that using positive tones and goal-oriented instructions works better than using negative constructs such as “do not” or “never use”. Such constructs tend to either confuse the model or shift its focus toward what we were trying to avoid. This led to a prompt that focuses on clear goal-oriented phrasing that describes what the model should do, not what it should avoid.

Afterwards, different prompting techniques were explored; zero-shot prompting \cite{brown_language_2020} was used as a baseline, chain of thought (CoT)  to encourage reasoning steps \cite{wei_chain--thought_2022}, and one-shot prompting \cite{brown_language_2020} to try and steer the model in the right direction. Each of the techniques were evaluated by the created metrics, which led to the conclusion that zero-shot prompting and subtle CoT work best for generating answers to student questions. According to our findings, this is because long and complicated prompts tend to confuse the model and lead to unexpected behaviors.

\section{Results}
First, we validate our custom LLM-as-a-Judge metric that is used to score how well an answer aligns with the teaching points covered in an expert-defined ground truth. We then use this validated scoring approach, together with a cost-per-request metric, to (i) establish a realistic human baseline, (ii) optimize the prompt, (iii) compare models within a single model family, and (iv) benchmark models across providers.

\subsection{Validity of automated scoring (LLM-as-a-Judge)}
\begin{figure}[H]
    \centering
    \includegraphics[width=\textwidth]{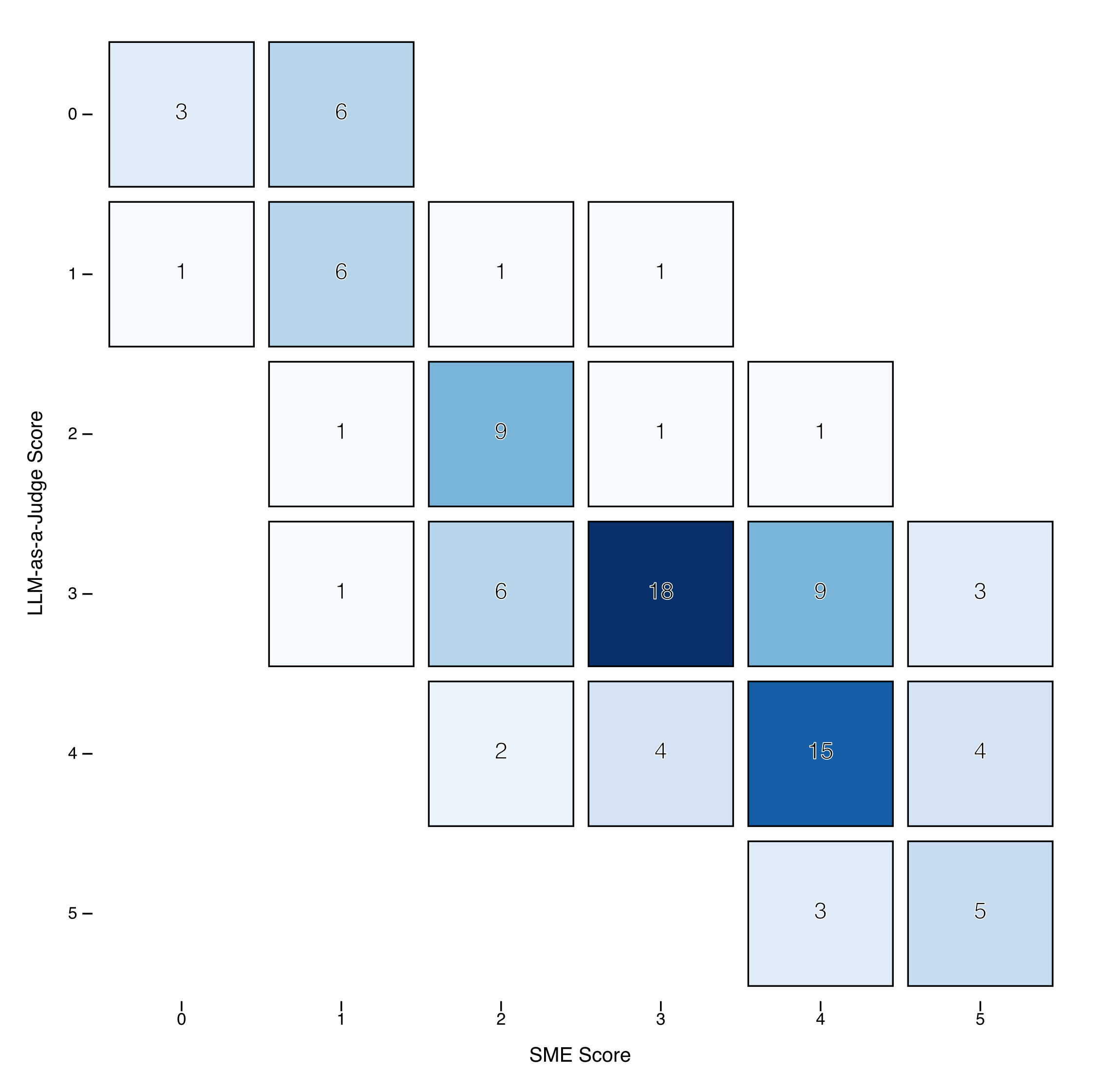}
    \caption{Heatmap showing the difference between SME-scores and the scores assigned by the LLM-as-a-Judge, where scores express the alignment between an actor’s answer and an expert’s reference answer.}
    \label{fig:figure3}
\end{figure}

Because answers to student questions are open-ended and can vary in phrasing and structure, we evaluate responses using a custom LLM-as-a-Judge: given an expert-authored ground truth answer as a reference, the judge assigns an ordinal score from 0–5 to an actor’s answer based on coverage of the core teaching points in the reference answer. Before using this score to evaluate prompts and models, we calibrated and validated the judge on a separate set of 100 responses that were independently scored by a subject-matter expert (SME) using the same scoring rubric. Averaged across three runs on this calibration set, the judge matched the SME score exactly in 55\% of cases, and 91.66\% of scores fell within ±1 point on the 0–5 scale. Agreement beyond chance was substantial (linear-weighted Cohen’s $\kappa_w = 0.655$), and rank/linear associations between judge and SME scores were strong (Pearson $r = 0.8185$; Spearman $\rho = 0.7982$; Kendall $\tau = 0.7067$). To visualise the alignment, we plotted the frequency of SME and LLM score pairs as a heatmap (Figure~\ref{fig:figure3}). High agreement between expert and automated judgments is indicated by a strong concentration of values along the diagonal. Full details of the rubric, judge prompt, and calibration procedure are provided in the supplementary material (judge prompt) and Appendix \ref{appendix:a} (calibration).

\subsection{Prompt engineering}
\begin{figure}[H]
    \centering
    \includegraphics[width=\textwidth]{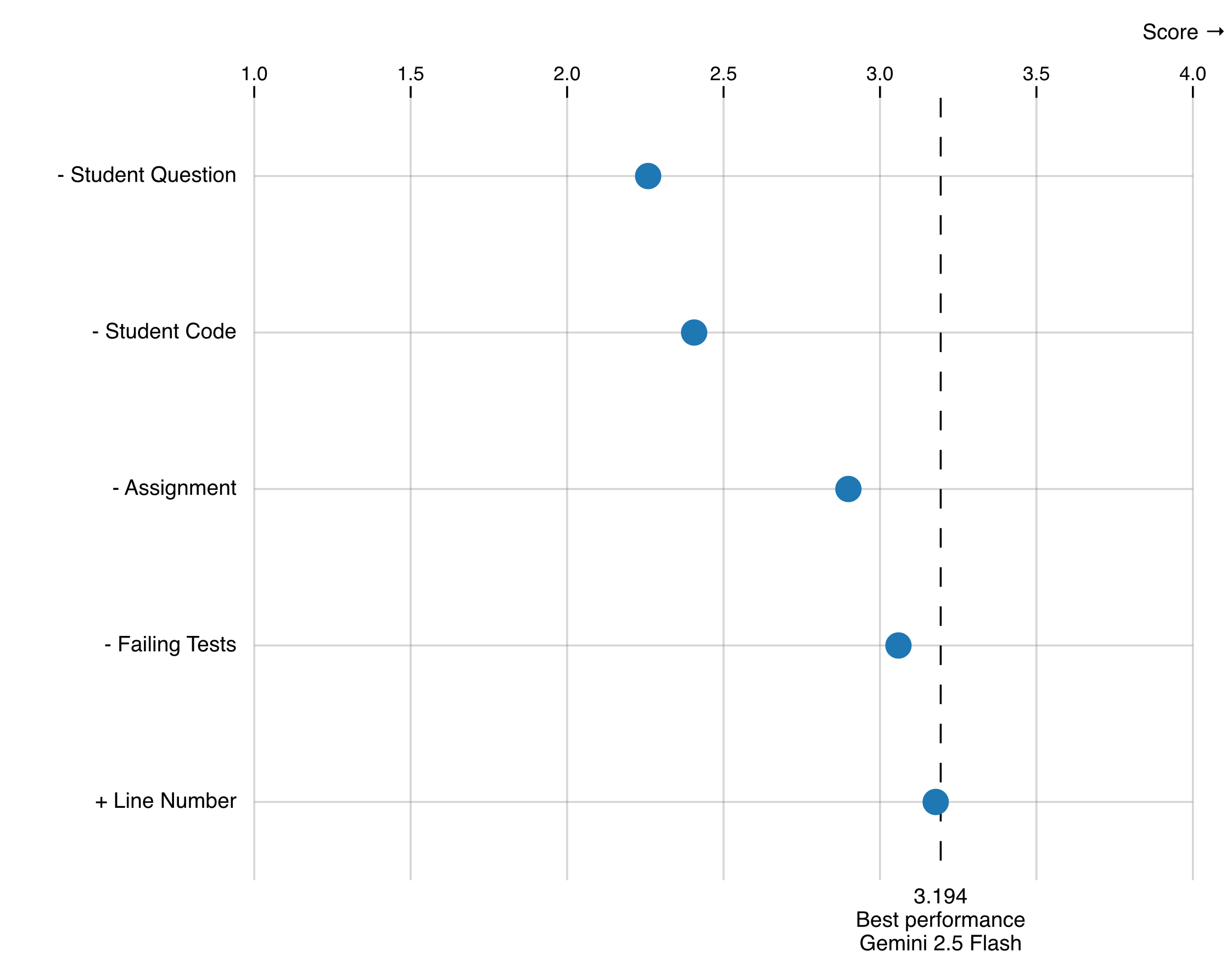}
    \caption{Q\&A task accuracy score of the Gemini 2.5 Flash model related to the input data provided to the model. The baseline is represented by the final and best performing prompt containing all input data except for the line number. `-` indicates a piece of data was left out of the baseline prompt, `+` indicates a piece of data was added.}
    \label{fig:figure4}
\end{figure}

After establishing the alignment of the LLM-as-a-Judge, we created a suitable prompt that can be used to evaluate the performance of different models. This prompt is engineered by testing different variations on a cheap but well-performing model. In our research, the Gemini 2.5 flash model was used to compare a huge amount of different prompt variations. These were easily evaluated on the input data from the programming course using the LLM-as-a-Judge metric. Variations in input data, data format and different prompting techniques were tested to ensure optimal performance. The final prompt uses the student’s question and code, the description of the assignment, failing tests and linting errors, the programming language and the students’ preferred natural language. Only the position of the students’ question (line number) had no impact on the performance of the model (Figure~\ref{fig:figure4}). XML-tags were used as delimiters and large bodies of text were converted to Markdown\cite{gruber_daring_nodate}. The used prompting technique is zero-shot prompting with chain-of-thought. The optimization process was highly iterative and exploratory, involving numerous rapid, ad-hoc tests rather than strict, linear progression. Consequently, a comprehensive performance breakdown for every prompt variation falls outside the scope of this research.

\subsection{Intra-family comparison}
\begin{figure}[H]
    \centering
    \includegraphics[width=\textwidth]{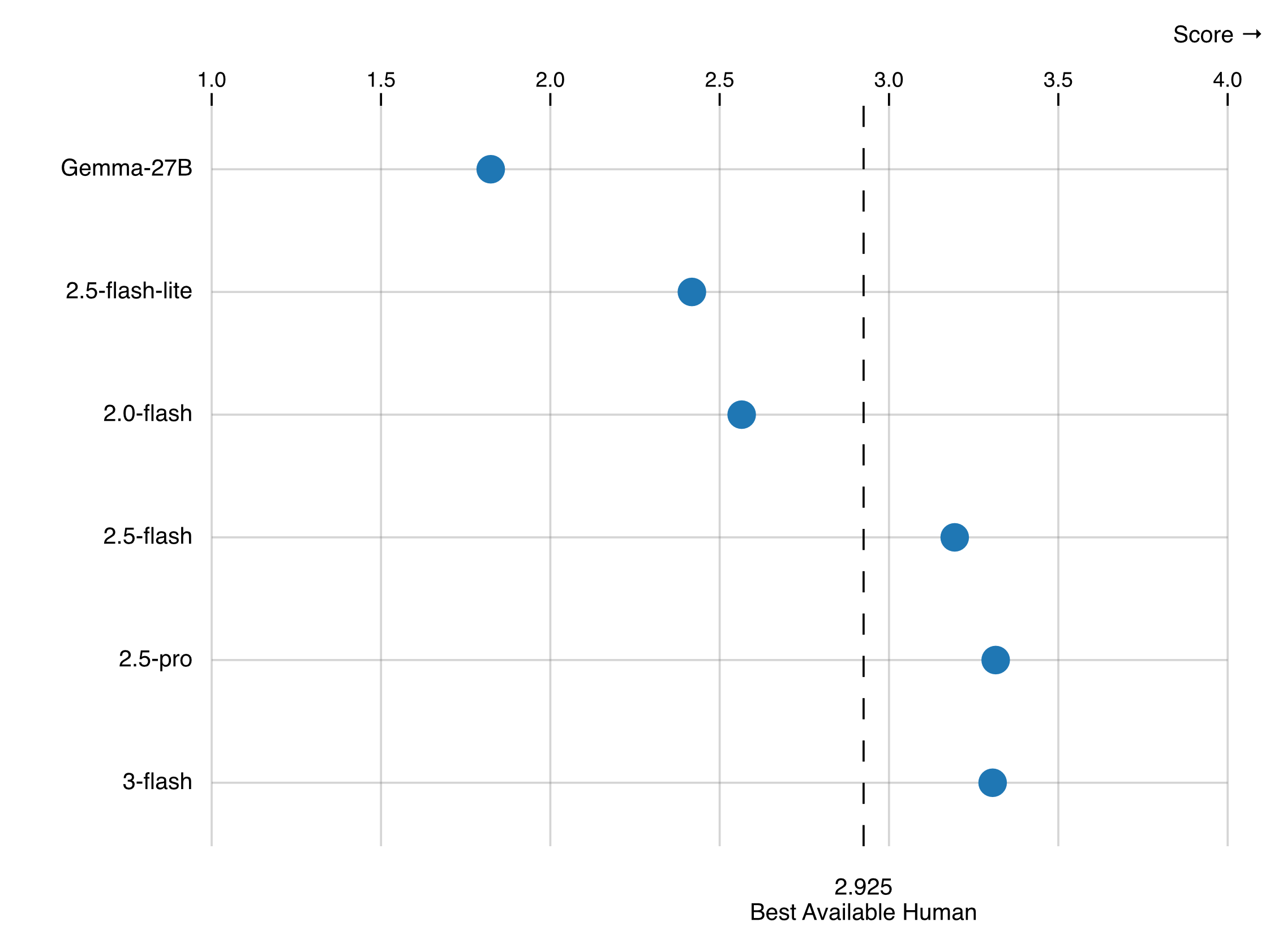}
    \caption{Q\&A task accuracy score of different models within the Gemini family performing the Q\&A task. Thinking was disabled on all models.}
    \label{fig:figure5}
\end{figure}

An LLM’s specific version and architecture has a significant impact on its performance. To evaluate this impact, multiple Gemini versions and variants were tested and compared against one another (Figure~\ref{fig:figure5}). The original answers provided by the educators (BAH) were used as a baseline, serving as a critical point of comparison for LLM performance, yielding a mean score of 2.925 out of 5. Among the tested models, Gemma-27B (a 2024 open-weights model that can be run locally) scored the lowest, with a mean of 1.82 out of 5, performing below the human baseline. Most modern models, however, surpass the human baseline, with Gemini 2.5 pro scoring 3.32 out of 5. These results suggest that modern models are capable of producing answers that not only align with expert expectations but also exceed the quality of educator responses generated in a realistic education context.

\begin{figure}[H]
    \centering
    \includegraphics[width=\textwidth]{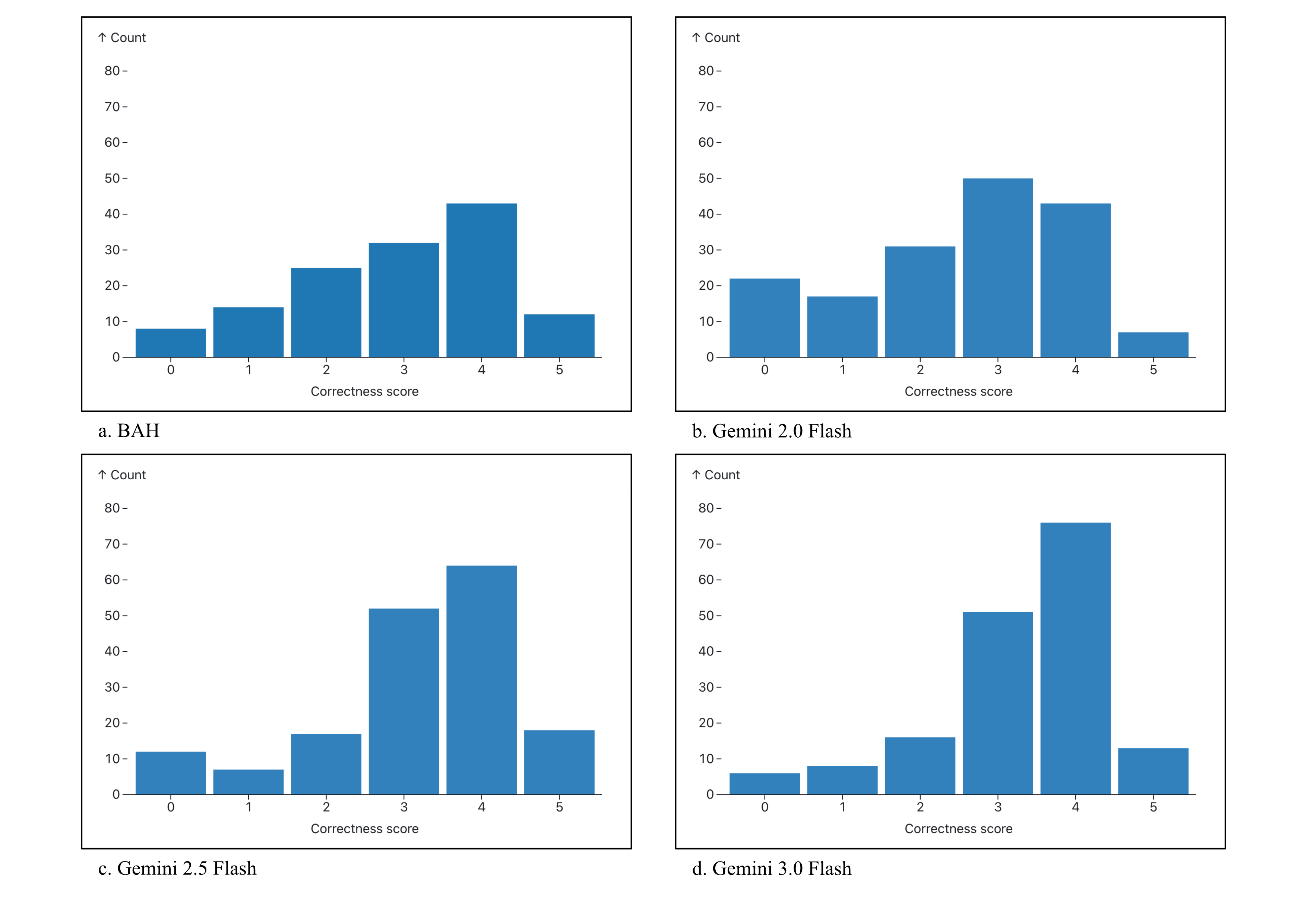}
    \caption{Score distribution of (a) the best available human, (b) Gemini 2.0 Flash, (c) 2.5 Flash without thinking and (d) 3.0 Flash with minimal thinking performing the Q\&A task. Graph (a) only contains data points that received a valid answer from an educator.}
    \label{fig:figure6}
\end{figure}

An analysis of the score distributions of the models within the Gemini family also shows a correlation between model version and obtained scores (Figure~\ref{fig:figure6}). Newer models have a stronger positive skew towards high scores compared to the older models. The newest model (Gemini 3 flash) obtains a score of 4 out of 5 or higher in 52.35\% of the cases. Smaller models, such as Gemini 2.0 Flash, only scored 4 out of 5 or higher in 29.4\% of the cases. Additionally, 82.35\% of the answers generated by Gemini 3 flash scored 3 out of 5 or higher, compared to 58.8\% of the answers generated by Gemini 2.0 Flash. Figure~\ref{fig:figure6} (a) contains the score distributions for the BAH (answer given by educators). Only 134 out of the 170 questions in the dataset have a valid answer provided by an educator. The scores are somewhere in between those of Gemini 2.0 Flash and Gemini 2.5 Flash: 41.05\% of answers scored 4 out of 5 or higher and 64.93\% scored 3 out of 5 or higher. This indicates that the most recent models perform at least as well as educators.

\subsection{Inter-family comparison}
\begin{figure}[H]
    \centering
    \includegraphics[width=\textwidth]{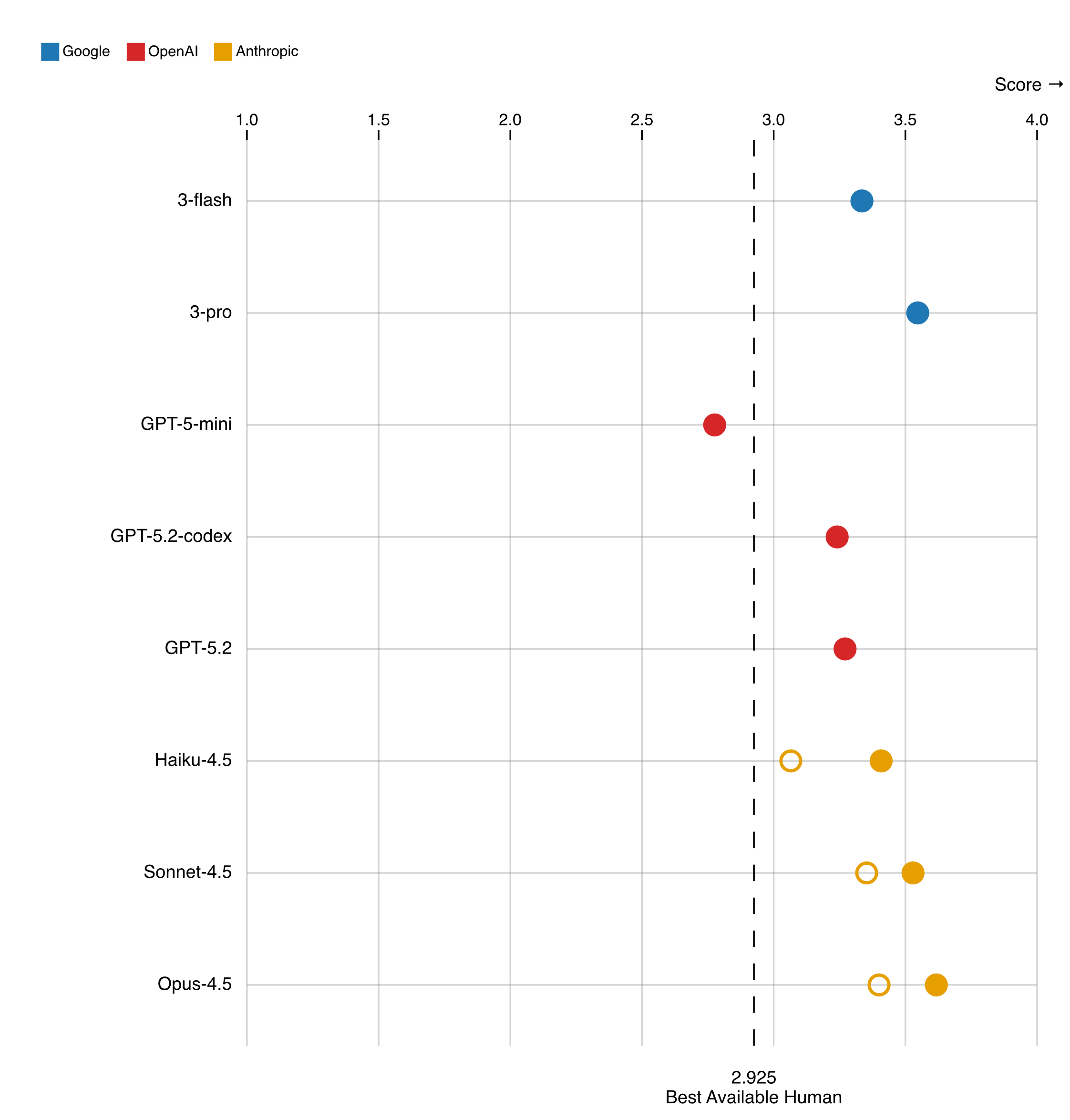}
    \caption{Q\&A task accuracy score of the tested models from Google, OpenAI and Anthropic. Solid dots represent performance with thinking enabled. Outlined dots represent performance with thinking disabled.}
    \label{fig:figure7}
\end{figure}

Leveraging the established prompt and metrics, we extended our evaluation to include models from different providers. Three popular families (Google's Gemini, OpenAI's GPT and Anthropic's Claude) were chosen and their most recent models were tested (Figure~\ref{fig:figure7}). Every run used the 170 student questions from the Q\&A evaluation dataset, the same prompt, and all other default settings of the model. All runs used temperature 0.2 unless the API imposed a different default; for Gemini 3 flash preview and Gemini 3 pro preview, temperature 1 was used due to technical limitations. Gemini models used dynamic thinking with all other default settings. Models from the OpenAI family used the default thinking effort (medium). The Claude models have thinking disabled by default, therefore, these models were run both with thinking enabled and disabled. When thinking was enabled for these models, the reasoning effort was set to medium.

With the exception of GPT 5 mini, all evaluated SOTA models outperformed the human baseline (2.925 out of 5). The models exhibit the same behavior as those within the Gemini family: flagship models consistently perform better. However, performance gains diminish as the benchmark becomes saturated. Newer models easily achieve scores close to four out of five, limiting the room for improvement. Moreover, because the task involves open-ended questions, achieving a perfect score is inherently difficult. As a result, additional model scaling yields smaller improvements, such that even the cheapest models perform very well on this task.

\begin{figure}[H]
    \centering
    \includegraphics[width=\textwidth]{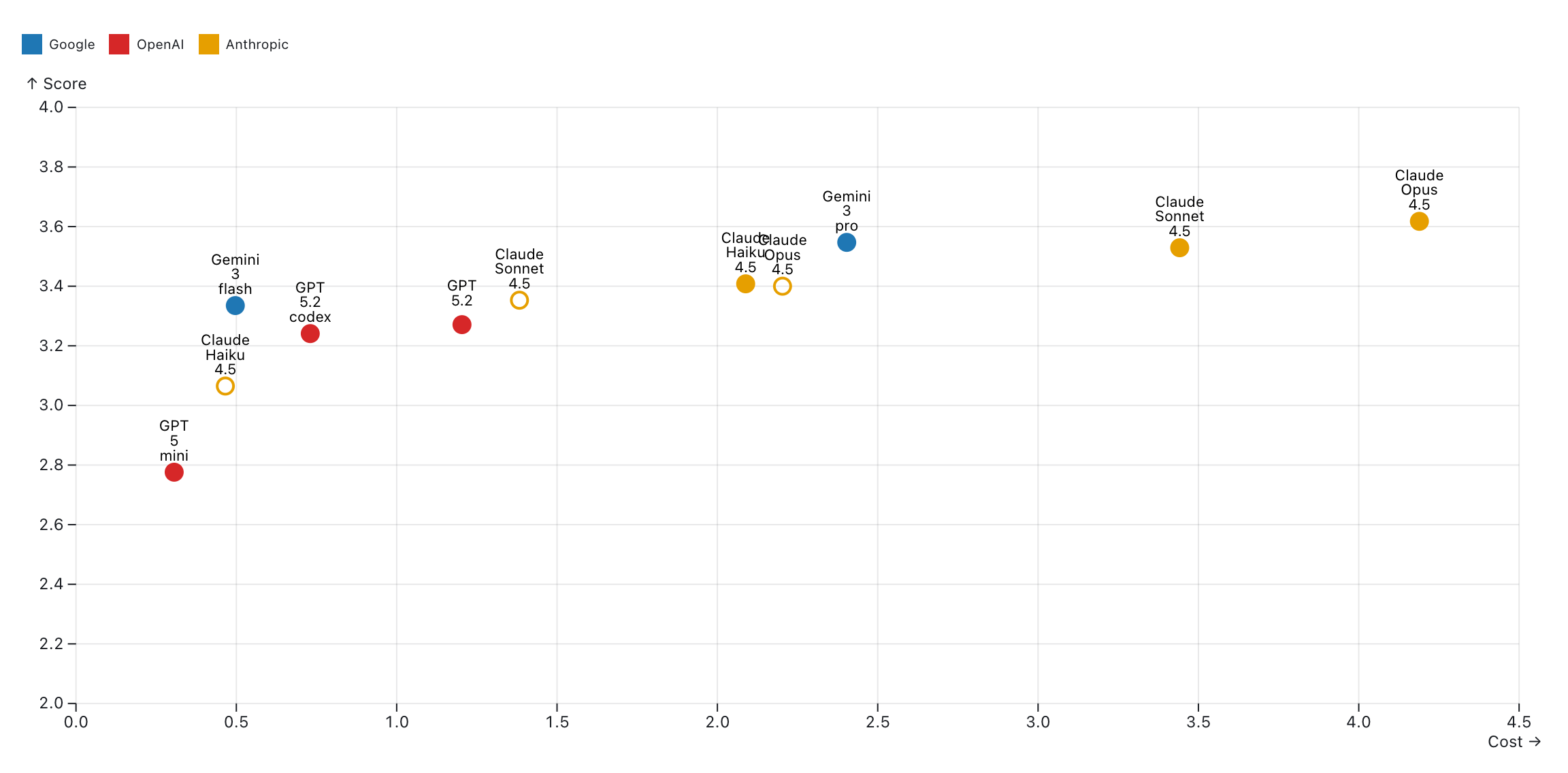}
    \caption{Q\&A task accuracy score of models from Google, OpenAI and Anthropic plotted against the average cost per request in USD cents. Models positioned toward the top-left represent the ideal balance of high accuracy and low cost. Solid dots represent performance with thinking enabled. Outlined dots represent performance with thinking disabled.}
    \label{fig:figure8}
\end{figure}

Moving towards the more expensive flagship models also comes with a bigger cost (Figure~\ref{fig:figure8}, Supplementary Fig. S.1). The average cost of a request to a model ranges from 0.306 USD cents for the most efficient model to 4.189 USD cents for the most expensive flagship model, a difference by an entire order of magnitude. One of the models with the best trade-off between performance and cost is Gemini 3 flash with a price of 0.497 USD cents per request and an accuracy score of 3.335 out of 5. If that model had been used during the 2023-2024 edition of the programming course (1140 questions), the total cost would have been 5.67 USD for one semester. Moreover, newer thinking models are more token efficient than previous generations (Supplementary Fig. S.2), reducing costs even further. However, if costs should be kept as low as possible, thinking can be disabled or thinking effort can be reduced. This is shown by the performance of the Anthropic models with thinking disabled. Their performance is lower than the performance of the models with thinking enabled, but their cost is also considerably lower. The cost of Claude Opus 4.5, for example, drops 47.41\% while performance only drops 6.03\%. As a result, most flagship models with or without thinking can be used to answer questions from students as they outperform the human baseline.

\section{Task-agnostic evaluation framework}
\begin{figure}[H]
    \centering
    \includegraphics[width=\textwidth]{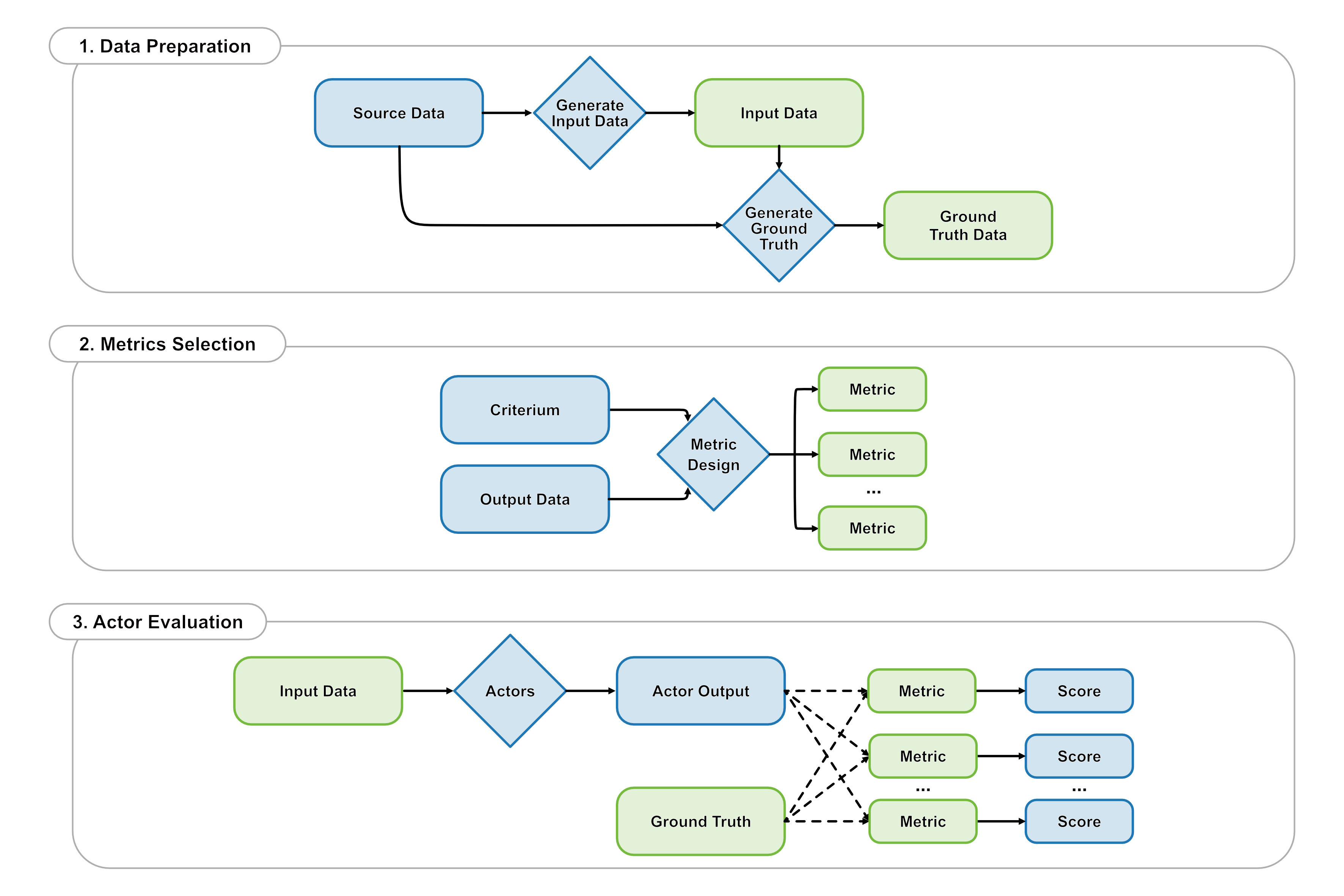}
    \caption{General framework to automatically evaluate the performance of different actors (prompt/model combinations, humans or a combination of LLM and human) in performing a given task.}
    \label{fig:figure9}
\end{figure}

Based on the research steps taken during the evaluation of the Q\&A task, we abstracted a task-agnostic evaluation framework (Figure~\ref{fig:figure9}). The framework focuses on the pre-deployment evaluation of different actors performing a certain task. These actors can be different prompt/model combinations, humans or agentic workflows. The framework comprises a preliminary feasibility study, followed by three sequential stages: Data Preparation, Metrics Selection and Actor Evaluation. During the data preparation stage, the evaluation dataset is created, consisting of the input data and ground truth. After compiling the evaluation dataset the success criteria are defined and matching metrics are selected or created. Given the input data, ground truth and metrics, a selection of actors is evaluated in the final stage. The input data is passed to each actor, resulting in an actor output. This output is combined with the ground truth and passed on to the predefined metrics. Each metric determines a score which can be used to compare or rank the different actors. Based on the gathered insights, the optimal actor can be selected for the given task. In what follows, we will discuss these steps in more detail.

\subsection{Feasibility Study}
Before automating the evaluation process, the feasibility of using an LLM for the targeted task should be evaluated. This does not only incorporate the ability of an LLM to perform this task but also the ethical and legal aspects related to the task at hand. For example, are you allowed to use an LLM to perform this task? Is it pedagogically acceptable? After these considerations, the capabilities of the LLMs can be tested. This process begins with curating a small, representative dataset and engineering a basic prompt. The LLM's outputs are then manually evaluated: if the prompt works acceptably, the evaluation process can start. If not, the prompt is iteratively refined as long as there are improvements. The results of this feasibility study inform the "go/no-go" decision for starting the evaluation process.

\subsection{Data Preparation}
For the Q\&A task, input data was derived from student questions stored in Dodona. We sampled 200 questions and manually filtered them to ensure GDPR compliance and to remove unrelated questions (e.g. questions about scores on tests). For each of the remaining 170 questions the ground truth was established by one of the supervising assistants of the course, resulting in the Q\&A evaluation dataset.

This process constitutes the first step in the evaluation framework. An evaluation dataset is curated, consisting of the input data (e.g. student questions and other context) and the ground truth (SME outputs to the input data). The input is sampled from raw source data, and is manually reviewed and edited to ensure optimal coverage of the source data. During the review process, crucial insights are gathered and missing samples are identified. Based on these insights, certain samples are added or removed from the input data. After the input data is selected, one or more SMEs start with the annotation of the input data to create the ground truth. The ground truth serves as the single source of truth for the task given the related input data. Ideally the ground truth is compiled by multiple experts independently, possibly followed by compiling a common ground truth among all experts. The evaluation dataset will be used to evaluate all current prompts, models and future implementations of the task, making this a vital and possibly time-consuming step in the framework.

\subsection{Metrics Selection}
For the Q\&A task, we identified pedagogical accuracy and cost-efficiency as the success criteria. In order to successfully complete the Q\&A task, the prompt/model combinations had to respond with an answer that aligned with the ground truth while keeping the cost as low as possible. These success criteria were translated into two metrics, a custom LLM-as-a-Judge that compares actor output against the ground truth and the average cost per request.

The metrics selection stage follows the same pattern: first the success criteria are defined. These criteria represent the qualities that the generated output should contain. Subsequently, appropriate metrics are chosen or created. These metrics will vary based on the success criteria, for example if the criterion is answers shorter than 100 words, the metric will be word count. However, if the criterion is a short answer that contains all information, the metric will have to take into account the amount of information and the word count. Regardless of the metric, the output should be closely examined to ensure that the metric aligns with human/expert evaluations. Discrepancies between the chosen metric and human evaluation could indicate that the metric is not fit for the task it needs to perform. For example, traditional machine learning metrics cannot judge sentiment or pedagogical relevance.

Our recommended approach when choosing metrics is as follows: Where possible, non-LLM-based metrics should be preferred, as they offer greater interpretability, reproducibility and lower susceptibility to (model-induced) bias. Traditional metrics such as BLEU \cite{papineni_bleu_2001}, ROUGE \cite{lin_rouge_2004}, METEOR \cite{banerjee_meteor_2005}, or exact match provide a reliable starting point. However, these metrics often fall short in tasks involving open-ended generation, where lexical overlap is a poor proxy for quality \cite{wang_can_2025}.

In cases where traditional metrics do not correlate with human judgment, general-purpose LLM-based evaluators developed and validated by third parties can be used. These models are typically used to assess qualities such as relevance, fluency, helpfulness and factual correctness. However, LLM-as-a-Judge systems do rely on prompt engineering and careful observations to produce consistent scores. Despite being flexible and general-purpose, they may introduce bias or lack transparency.

Only when both the traditional and existing LLM-based evaluators fail, a custom LLM-as-a-Judge should be developed. This custom judge should be manually evaluated and compared to expert-level human scoring. Our automated evaluation framework for prompt and model selection can be used to make sure the judge performs as expected by iteratively improving the prompt used by the judge.

\subsection{Actor Evaluation and Selection}
The final step in our Q\&A research was testing the best available human output, evaluating different prompts, performing an intra-family comparison and ending with a comparison of models from different providers. Based on the accuracy scores obtained by all of these evaluations, we were able to conclude that LLMs outperform human educators on the Q\&A task, making them suitable for deployment as draft-generating assistants within a teacher-in-the-loop workflow.

This repeated evaluation is the final step in our framework. The input data is given to each actor (prompt, model, a human annotator, …) which results in an actor output. This output and the ground truth linked to the input data are handed to each of the metrics created in the second stage of the framework. This results in multiple scores that can be used to make a ranking, plot or overview. Subsequently, the data can be used to draw conclusions about the different actors performing the task.

\section{Discussion}
This study set out to explore the potential of LLMs to support educators in CS1 programming courses by generating answers to student questions, guided by three research questions.

\subsection{RQ1: To what extent can LLMs generate pedagogically appropriate answers to student questions in a CS1 programming course?}
Our findings indicate that current LLMs are capable of answering student questions in a CS1 programming course, with models like Gemini 3 flash surpassing the quality of typical educator responses. The "best available human" (BAH) baseline, representing the quality of time-constrained educator feedback, achieved a mean score of 2.92 on our 0-to-5 scale. In comparison, most models like Gemini 3 flash surpassed this score with 3.335 and higher. This superior performance suggests that modern LLMs are capable of generating answers that not only align with expert answers but also exceed the quality of typical educator responses. The superior performance of LLMs might be attributed to their ability to quickly generate detailed and comprehensive explanations, whereas educators may be limited by time constraints, leading to less detailed feedback. 

However, it is important to acknowledge that some LLM-generated information, while extensive, might be providing overly detailed or inaccurate information, giving away solutions or hallucinating, which can mislead the student and cause a loss of trust. To mitigate these issues, a Q\&A tool was implemented in Dodona following a teacher-in-the-loop approach. An LLM generates a draft answer that is then reviewed and edited by an educator before being sent to the student who had asked the question. This human-augmented LLM tool is intended to produce more accurate answers that reach the student faster. The best performing prompt from this research combined with the Gemini 2.5 flash model is currently in use. Before deployment, the tool was extensively tested on the Q\&A benchmark, and future models and improved prompts can be tested on the same benchmark. Future research will investigate how the "teacher-in-the-loop" method affects accuracy, educational impact, and responsiveness.

\subsection{RQ2: How can we establish a reproducible, scientific process for developing and evaluating LLM-based (educational) tools?}
The evaluation of the Q\&A task serves as a proof-of-concept for RQ2, demonstrating that we can establish a reproducible, scientific process for developing (educational) LLM tools. By defining a benchmark and metrics that are specifically representative for the task at hand and the goals we try to accomplish, we established a method to rigorously quantify the performance and cost of LLMs. This approach allowed us to assess the effectiveness of different prompt/model combinations and even the BAH, providing a concrete overview of the current LLM capabilities to perform the (educational) task.

Beyond pre-deployment assessment, this benchmark provides the necessary infrastructure to future-proof the development process. It enables systematic re-evaluation of the tools as the LLM landscape evolves. The influence of new models, updated prompts or the introduction of new technologies is quantifiable through our benchmark. Consequently, this shifts the development of (educational) LLM-tools from ad-hoc implementation with post-deployment evaluation to a future-proof, quantifiable, scalable and reproducible pre-deployment process.

\subsection{RQ3: What reusable workflow or set of principles can be distilled for designing and evaluating similar tools across domains and tasks?}
Based on the design and evaluation of our Q\&A task, we were able to abstract an evaluation framework that provides us with a reliable way of quantifying the performance of different actors (referring to prompt/model combinations, humans, or a hybrid of both) performing a task (Figure~\ref{fig:figure9}). This framework moves the development of LLM-based tools from a trial-and-error approach to a structured, task-agnostic pipeline governed by three core principles: (i) gather insights, (ii) automate the evaluation and (iii) future-proof the evaluation.

The first principle, gathering insights, is rooted in the early stages of the framework. During a preliminary feasibility study, data preparation and metric selection, researchers are forced to deeply engage with the source data and task requirements. By curating the evaluation dataset and success criteria upfront, crucial insights into the data’s nuances and pitfalls are gained. By defining the goals and identifying potential issues early on, the overall focus of the task is clarified, which reduces errors and workload in later stages.

Building upon the initial insights, the focus shifts towards automating the evaluation. The core strength of the proposed framework lies in its ability to translate success criteria into automated, quantifiable metrics. By thoughtfully selecting or designing automated metrics, the bottleneck of labor-intensive manual evaluation is eliminated. Automation does not only reduce time and the required financial resources, but also provides quantifiable scores that are necessary to reliably evaluate and compare different actors.

With the automatic metrics and evaluation dataset in place, we can create a reusable testing pipeline (actor evaluation phase). As the LLM landscape evolves, new models, prompts or agentic workflows can be used as actors in the evaluation pipeline and evaluated instantly. There is no need for new datasets, manual annotation or new metrics. LLM tools can be continuously re-evaluated to maintain state-of-the-art performance.

The versatility of the task-agnostic framework is demonstrated by its application to our custom evaluation metric. The custom LLM-as-a-Judge was developed and refined using the same framework and based on the same principles (Appendix A). By treating the judge as an actor performing a specific task (evaluating pedagogical accuracy), we could quantify its alignment with human expert scoring. Furthermore, by building an automated, future-proof pipeline for the evaluation of the LLM-as-a-Judge, the judge can be refined and re-evaluated with new model/prompt combinations.

\section{Limitations}
While this study shows the capabilities of LLMs in an educational Q\&A setting, several limitations must be acknowledged. First, the data used in this study originates from one CS1 programming course taught in Python. Consequently, the performance of the tested models may vary in upper-level courses or courses that use different programming languages and paradigms. Second, the annotations constructed for the 170 selected student questions were annotated by a single expert. While this approach ensures a pedagogically accurate answer, the response may not encompass all possible solutions or variations in teaching style. Although the LLM-as-a-Judge compensates for this by focusing on the core teaching points, the judge remains an automated proxy for manual expert evaluation. Finally, the performance of the LLMs used in this study is a snapshot in time, with the rapid evolution of generative AI re-evaluation using our proposed framework will be necessary.

\section{Conclusion}
Our research demonstrates that LLMs show promising performance in answering CS1 student questions, with models often surpassing time-constrained educator responses. This suggests that LLMs can serve as effective support tools in educational contexts, reducing workload and improving responsiveness. However, occasional inaccuracies highlight the importance of human oversight to ensure correctness and maintain trust in learning environments. To address this, we implemented the Q\&A tool in Dodona with a teacher-in-the-loop approach. This reduces the risk of misinformation, increases answer quality and reduces educator workload.

Additionally, the research for the Q\&A task provides us with a reproducible, scientific process for developing educational LLM tools. By creating task-specific benchmarks and metrics before deployment of the tool, we quantified LLM performance of various prompt/model combinations and the BAH. From this process, we distilled a task-agnostic evaluation framework, shifting the development of educational AI tools from ad-hoc, post-deployment assessment to a quantifiable, scalable, and reproducible validation process. This provides a future-proof pipeline for the systematic evaluation of new models and prompts as the generative AI landscape evolves.

Ultimately, our research highlights the potential of LLMs as powerful educational aids when integrated responsibly. By combining human oversight with pre-deployment evaluation and by providing a framework to perform this evaluation, we pave the way for safer, more effective and scalable adoption of LLMs in education.

\section*{Acknowledgements}
TVM, BM and PD acknowledges funding by Research Foundation - Flanders (FWO) for ELIXIR Belgium [I002819N].

\bibliographystyle{plain}
\nocite{*}
\bibliography{bibliography}

\appendix
\renewcommand\thefigure{\thesection.\arabic{figure}}
\counterwithin{figure}{section}

\section{Design of a custom LLM-as-a-Judge to compare generated answers to a ground truth}
\label{appendix:a}
The automation of the comparison between SME answers (ground truth) and others (generated answers) is based on the similarity between the ground truth and those other answers. These “other” answers can be generated by an LLM or written by educators. The similarity between two answers is based on the correctness and completeness of the information. In the context of drafting answers to programming questions, correctness and completeness can be interpreted as conveying the same teaching points without adding or missing information. It is also important to note that there is not a single perfect answer to a programming question. The ground truth answer represents one possible answer to the question asked by the student. This ambiguity makes it difficult to use an already existing metric to measure the correctness of an answer compared to a ground truth dataset. To solve this issue, we introduced a custom LLM-as-a-Judge metric. The metric assigns a correctness score to every answer and corresponding ground truth item, reflecting how accurate the answer is relative to the ground truth. To make sure this judge is correctly aligned with the opinion of an SME we used the evaluation framework. The task is to align the judge with an SME by automatically evaluating the difference in scores between them.

\subsection{Feasibility Study}
The metric selection phase of the Q\&A task showed that it is not feasible to use conventional NLP metrics to determine the similarity of generated answers and ground truth items. To automate the evaluation for the Q\&A task, we used LLM-as-a-Judge which has been proven to be a good replacement for manual evaluation. In our case, initial tests with items from the ground truth dataset and answers from intermediary runs with other metrics show that LLMs are able to give scores that align with those given by an SME. The feasibility study uncovered that the challenge in prompting the model for optimal alignment is hidden in the definition of `similarity`. Translating the criteria of the SME into a scoring rubric or prompt that can be provided to the model is the main challenge.

\subsection{Data Preparation}
During the data preparation stage the benchmark is established to align the custom LLM-as-a-Judge with an SME. Following the insights from the feasibility study, this stage focuses on creating a scoring rubric and gathering and annotating the necessary data to align the judge. A ground truth dataset is constructed, wherein similarity scores are assigned by an SME. This dataset, used for the alignment, is created by combining data from prior runs utilizing alternative metrics (Q\&A task) with the existing ground truth dataset from the Q\&A task. Each data point in the consolidated dataset is assigned a score between 0 and 5 by an SME.

\subsubsection{Dataset}
Based on insights gathered in the feasibility study and previous runs with different metrics during the Q\&A task, a ground truth dataset was constructed to align the LLM-as-a-Judge with the scores of an SME. This ground truth dataset is based on 100 samples from two different runs of the Q\&A task with other metrics. Fifty of those samples were taken from a run without limitations on the word count, the LLM was not hinted towards short answers nor was a word count specified. The other fifty samples were taken from a run where the LLM was instructed to limit the answers to 100 words unless more details were essential. The model was also instructed to ‘keep the answers short and focused, avoiding unnecessary details or filler’. This dataset represents two relevant scenarios that occurred in previous runs of the Q\&A task. The first scenario is characterized by lengthy explanations generated by the LLMs. These responses, while thorough, often include extra information that is not necessary to help the student. This verbosity can hinder or confuse the student. The second scenario involves the generation of concise short answers that, while efficient, could suffer from incompleteness. The LLM leaves out essential details or skips some of the necessary teaching points. Both scenarios are undesirable and result in a lower score.

\begin{figure}[H]
    \centering
    \includegraphics[width=\textwidth]{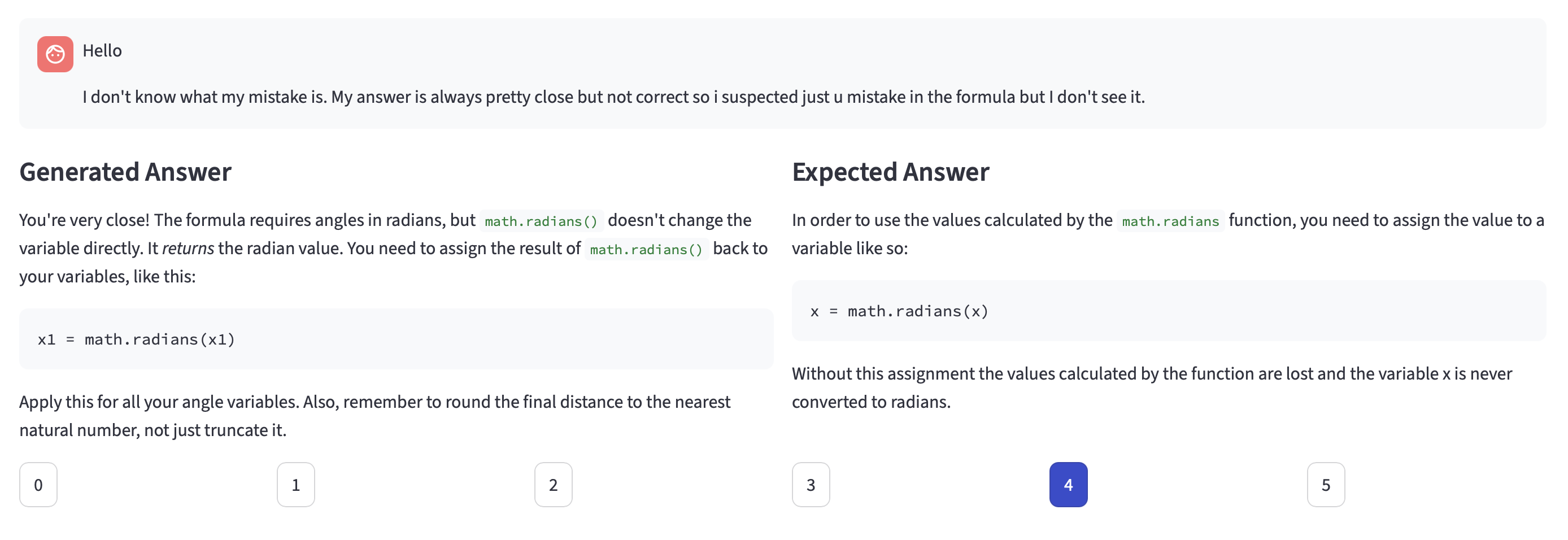}
    \caption{SME view when grading answers for the ground truth dataset of the LLM-as-a-Judge. The SME sees the generated and expected answer and gives a score between 0 and 5 based on a predefined score rubric.}
    \label{fig:figurea1}
\end{figure}

\subsubsection{Score Rubric}
The LLM-as-a-Judge metric focuses on the ‘core teaching points’ of the generated (LLM-answer) and expected answer (SME-answer). The task of the judge is to score generated answers based on the similarity of the core information that is presented by that answer. A scoring rubric is provided to the model to provide accurate descriptions of what each score represents. The metric focuses on the teaching points and the ability of the model to recognize the core information the student needs without overly relying on the reference answer. An answer that contains a solution that uses a different approach but has the same core meaning (solves the same issue) is considered correct. Complete matches result in a score of 5, a complete mismatch is assigned a score of 0. The SME uses this score rubric to give a score to each item in the ground truth dataset (Figure \ref{fig:figurea1}). This score is later used to align the LLM-as-a-Judge with the expert.

\subsection{Metric Selection}
To evaluate the scores given by the LLM-as-a-Judge a simple metric can be used. The difference between the SME and judge scores gives an indication of how well aligned they are. The goal is to minimize this difference for every item in the dataset. Metrics such as Cohen's (weighted) kappa \cite{cohen_weighted_1968} can be used to indicate the current alignment. Visual representations such as heatmaps can help an expert to determine what needs to change and can be automatically generated.

\subsection{Actor Evaluation}
\begin{figure}[H]
    \centering
    \includegraphics[width=\textwidth]{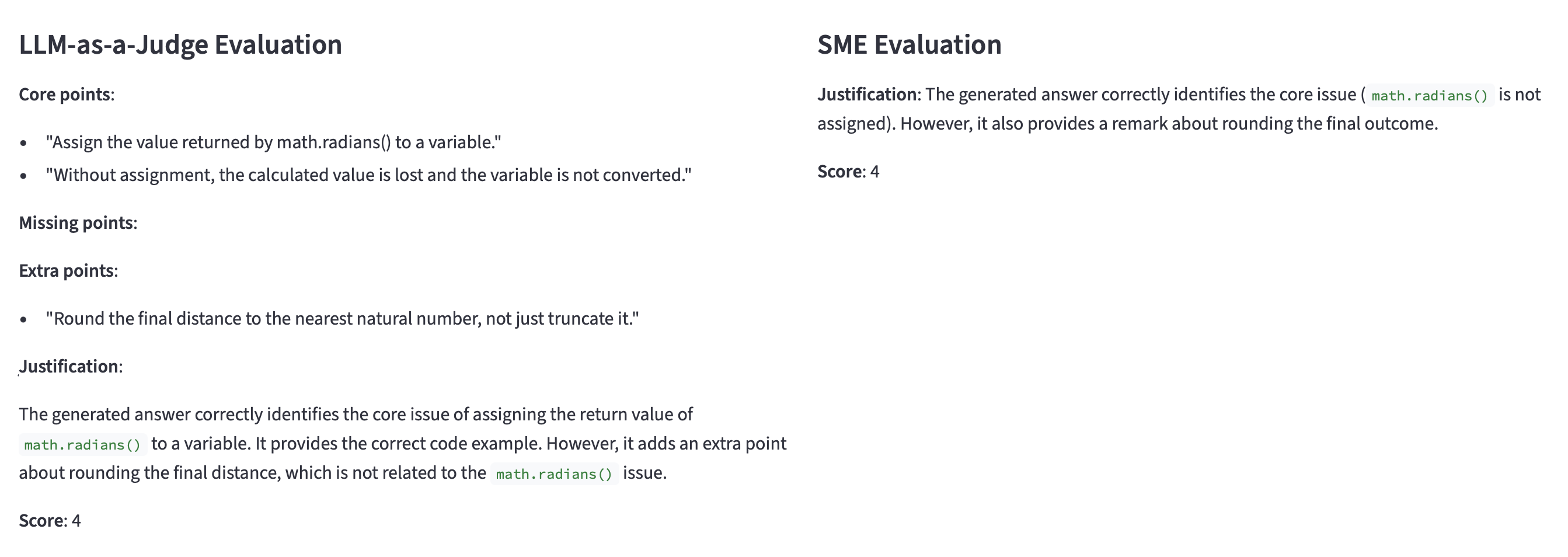}
    \caption{Output of the LLM-as-a-Judge and the score and justification of the SME for the radian example}
    \label{fig:figurea2}
\end{figure}

Lastly, the judge is aligned with the SME scores during the actor evaluation stage. Different prompting strategies are explored and each of the resulting score distributions is carefully studied to ensure optimal alignment. The wording of the prompt and scoring rubric are carefully adjusted to steer the judge towards the scores provided by the SME. Heatmaps are constructed and the explanations given by the judge are inspected (Figure \ref{fig:figurea2}). The necessary data is selected based on the highest scoring alignment. E.g. prompts are created where the student’s question or the identified issue provided by the LLM were included. This refinement process is repeated until no significant positive changes are detected.

\clearpage


\section*{Supplementary material}

# Supplementary Material

## Submission example (json)

The following JSON object is a sample entry from the input dataset, corresponding to the "radians example" shown in Fig. 1 of the main article.

```
{
    "submission_id": 1,
    "line_number": 0,
    "user_language": "English",
    "programming_language": "Python",
    "question": "Hello

I don't know what my mistake is. My answer is always pretty close but not correct so i suspected just u mistake in the formula but I don't see it.",
    "code": "import math
#step1: hoeken ingeven
x1= float(input())
y1= float(input())
x2= float(input())
y2= float(input())
#step2 hoeken omzetten in radialen
math.radians(x1)
math.radians(y1)
math.radians(x2)
math.radians(y2)
#step3: radius of th earth
r = 6371
#step3: calculating the great-circle distance
d =r*math.acos((math.sin(x1)*math.sin(x2))+(math.cos(x1)*math.cos(x2)*math.cos(y1-y2)))
# step 4: show the result on the screen
print(f'The great-circle distance is {int(d)} km.')",
    "exercise_question": "A great circle of a sphere is the intersection of the sphere and a plane that passes through the center point of the sphere. A great circle is the largest circle that can be drawn on any given sphere. Any diameter of any great circle coincides with a diameter of the sphere, and therefore all great circles have the same circumference and have the same center as the sphere.

For most pairs of distinct points on the surface of a sphere, there is a unique great circle through the two points. The exception is a pair of antipodal points, for which there are infinitely many great circles. The shortest distance between two points on the surface of a sphere \u2014 measured along the surface of the sphere \u2014 is always along a great circle of the sphere, and is therefore called the **great-circle distance**.

A great circle divides a sphere into two equal hemispheres.
```

*A transatlantic ship, for example, does not travel from Southampton, England to New York, USA alongside an east-west latitude \u2014 which appears to be the shortest route on many world maps (e.g. using Mercator projection) \u2014 but over the great circle that runs through both places. This great circle runs relatively high north over the Atlantic Ocean. The map below initially shows the part of the great circle that forms the shortest distance between Southampton and New York. You can drag the end points at will.*

*The great-circle distance $$d$$ between two points on a sphere can be calculated using the formula: $$d = r \\cdot \\arccos(\\sin(x_1) \\cdot \\sin(x_2) + \\cos(x_1)\\cdot \\cos(x_2) \\cdot \\cos(y_1 - y_2))$$ where $$(x_1, y_1)$$ and $$(x_2, y_2)$$ are the longitude and latitude of both points (given in decimal degrees) and $$r$$ is the radius of the sphere on which the distance is to be calculated.*

*In this assignment we assume that Earth is spherical with radius $$r = 6371\\mbox{ km}$$. Despite the fact Earth is more like a flattened spheroid rather than a perfect sphere, the formula for the great-circle distance still gives an approximation that is correct up to $$0.5\\%$$.*

*### Input*

*The input consists of 4 numbers $$x_1, y_1, x_2, y_2 \\in \\mathbb{R}$$, each on a separate line. The longitude-latitude pairs $$(x_1, y_1)$$ and $$(x_2, y_2)$$ represent the positions of two points on Earth, where longitudes and latitudes are expressed in degrees.*

*### Output*

*A description that indicates the great-circle distance (in kilometers) between the two given points on Earth. The great-circle distance must be rounded to the nearest natural number.*

*### Example*

**Input:**
```
    48.87
    -2.33
    37.80
    122.40
```

**Output:**```
    The great-circle distance is 8948 km.
```
*",
    "educator_answer": "These statements have no effect. They only compute the angles in radians from the angles in degrees, but never do something with the result of the computations. You need to reassign the results to the same variables that are then used further down in the program.

````Python
x1 = math.radians(x1)
…
````",
    "testcases": {
        "accepted": false,
        "status": "wrong",
        "description": "50 tests failed",
        "annotations": [
            {
                "column": 0,
                "externalUrl":
"https://pylint.pycqa.org/en/latest/messages/convention/missing-final-newline.html",
                "row": 16,
                "text": "Final newline missing",
                "type": "info"
            }
        ],
        "tabs": [
            {
                "contexts": [
                    {
                        "accepted": false,
                        "testcases": [
                            {
                                "accepted": false,
                                "tests": [
                                    {
                                        "accepted": false,
                                        "generated": "The great-circle distance is 9981 km.\n",
                                        "expected": "The great-circle distance is 8948 km.\n",
                                        "channel": "stdout"
                                    }
                                ],
                                "description": "48.87
-2.33
37.80
122.40
"
                            }
                        ]
                    }
                ],
                "description": "correctness"
            }
        ]
    }
}

# Judge prompt

The following text shows the prompt-template that is used to instruct the LLM-as-a-Judge. The prompt consists of general instructions, a scoring rubric, formatting information and specific instructions to perform the task.

---

You are an alignment judge for a programming course.
You will receive two answers:
- Expected Answer: the instructor's reference solution.
- Generated Answer: AI generated reply to a student.

Your task is to compare these two answers only. You do not see the original student question.
Judge whether the Generated Answer conveys the same core teaching points as the Expected Answer.
If the same core issue is addressed but a slightly different solution is proposed, that is acceptable.

Important:
- A score of 3 or higher means that the generated answer is acceptable to show to a student. This means that the core teaching points are covered, the student is able to solve the issue with the provided information without introducing more mistakes (no extra harmful information).
- A score of 2 or lower indicates that either more than one key point is missing, or that the generated answer contains a lot of extra information.
- Extra tips in the reference answer are not essential (not a core issue), but give additional information that may help the student.
- Don't be afraid to give the maximum score of 5 if the answers are very similar.

———

Scoring rubric (0–5)
- 5 (Complete alignment): Covers all information in the Expected Answer or suggests an equivalent solution.
- 4 (High alignment): Covers all core ideas in the Expected Answer but a detail is added or missing.
- 3 (Adequate alignment): Main ideas from the Expected Answer are present but the answer adds a few extra details or one significant step/core idea is missing.
- 2 (Low alignment): Partial overlap; answer is too verbose and contains unnecessary details or identifies unrelated core issues or is missing two or more key points.
- 1 (Poor alignment): Very little overlap; most key points are missing, or the answer is mostly irrelevant/misleading or contains incorrect information.
- 0 (Complete misalignment): Entirely unrelated, incorrect, or misleading; contains none of the Expected Answer's key points.

———

Output Format

Always return your evaluation in this YAML schema:

core_points: [ "<points from Expected answer>" ]
missing_points: [ "<points from Expected that are missing>" ]
extra_points: [ "<unnecessary or misleading additions>" ]
justification: "<2–4 sentences explaining summary of comparison>"
Score: <0-5>

———

Instructions
1. Identify the core teaching points in the Expected Answer.
2. Compare them against the Generated Answer.
3. Mark what is covered, missing, or unnecessarily added.
4. Assign a score using the rubric above.
5. Fill in the YAML output.

<Expected Answer>
{expected_output}
</Expected Answer>

<Generated Answer>
{output}
</Generated Answer>

# Flagship model statistics

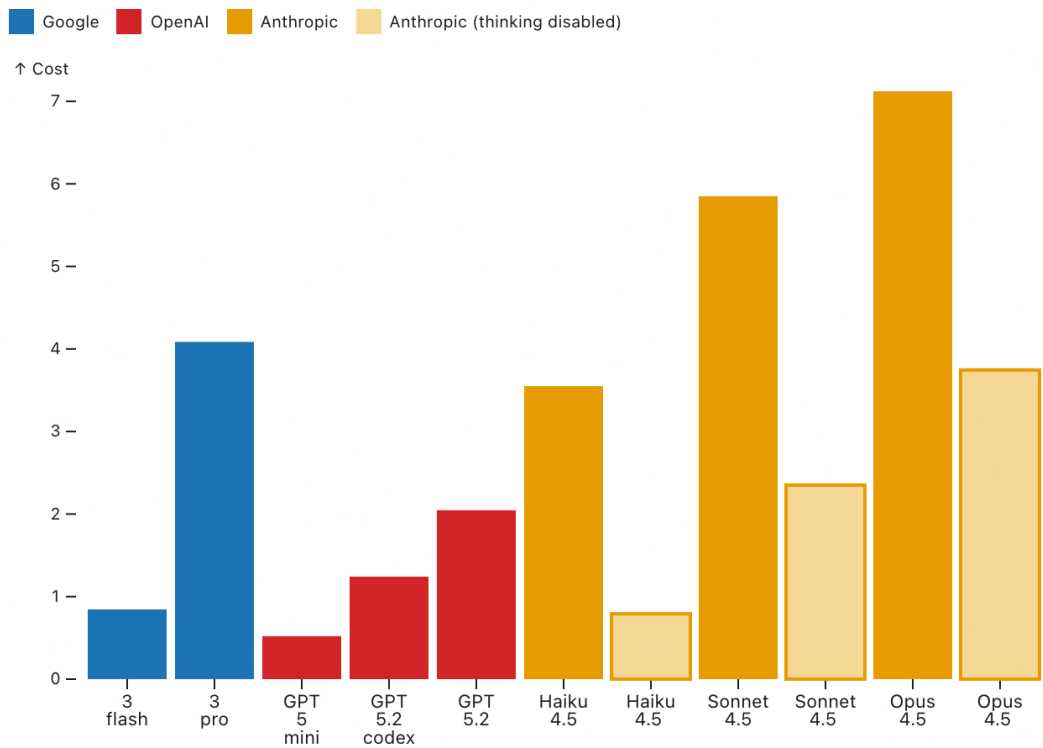

**Fig. S.1** Total cost for each actor during one run of the Q&A actor evaluation.

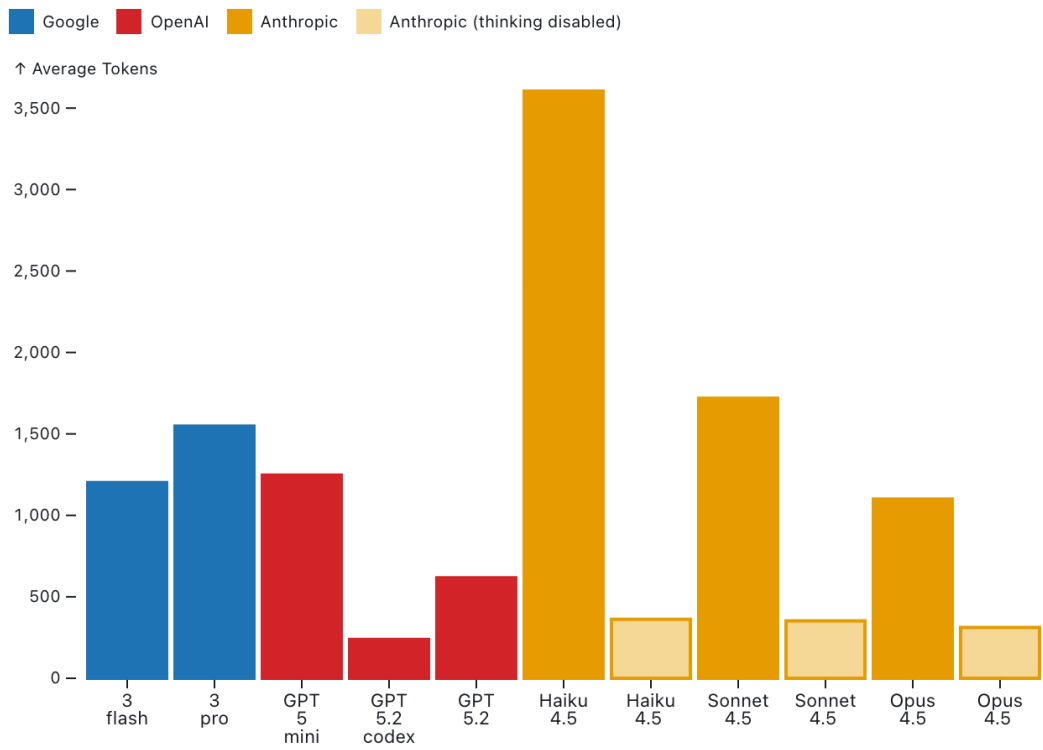

**Fig. S.2** Average token usage per request of the tested models during one run of the Q&A actor evaluation.



\end{document}